\documentclass[10pt,twocolumn,letterpaper]{article}

\usepackage{wacv}

\usepackage{amsmath}
\usepackage{amssymb}
\usepackage{booktabs}
\usepackage{graphicx}
\usepackage{import}
\usepackage{makecell}
\usepackage{multirow}
\usepackage{url}
\usepackage{threeparttable}
\usepackage[dvipsnames]{xcolor}
\usepackage{caption}
\usepackage{wrapfig}
\usepackage{xspace}
\usepackage[pagebackref,breaklinks,colorlinks]{hyperref}

\usepackage[capitalize]{cleveref}

\begin{document}

\crefname{section}{Sec.}{Secs.}
\Crefname{section}{Section}{Sections}
\Crefname{table}{Table}{Tables}
\crefname{table}{Tab.}{Tabs.}

\def\wacvPaperID{1099} 
\def\confName{WACV}
\def\confYear{2024}


\title{MACP: Efficient Model Adaptation for Cooperative Perception}

\author{%
  Yunsheng Ma$^1$\thanks{Equal Contribution}, Juanwu Lu$^1$\footnotemark[1], Can Cui$^1$, Sicheng Zhao$^2$, Xu Cao$^{3,5}$, Wenqian Ye$^{4,5}$, Ziran Wang$^1$ \\
  $^{1}$ Purdue University, West Lafayette, IN, USA\\
  $^{2}$ Tsinghua University, Beijing, China \\
  $^{3}$ University of Illinois Urbana-Champaign, Champaign, IL, USA\\
  $^{4}$ University of Virginia, Charlottesville, VA, USA\\ 
  $^{5}$ PediaMed AI, Shenzhen, China\\
  \texttt{\{yunsheng,juanwu,cancui,ziran\}@purdue.edu} \\
  \texttt{schzhao@tsinghua.edu.cn, xucao2@illinois.edu, wenqian@virginia.edu}
}
\maketitle

\begin{abstract}
Vehicle-to-vehicle (V2V) communications have greatly enhanced the perception capabilities of connected and automated vehicles (CAVs) by enabling information sharing to ``see through the occlusions", resulting in significant performance improvements. However, developing and training complex multi-agent perception models from scratch can be expensive and unnecessary when existing single-agent models show remarkable generalization capabilities. In this paper, we propose a new framework termed MACP, which equips a single-agent pre-trained model with cooperation capabilities. We approach this objective by identifying the key challenges of shifting from single-agent to cooperative settings, adapting the model by freezing most of its parameters and adding a few lightweight modules. We demonstrate in our experiments that the proposed framework can effectively utilize cooperative observations and outperform other state-of-the-art approaches in both simulated and real-world cooperative perception benchmarks while requiring substantially fewer tunable parameters with reduced communication costs. Our source code is available at \url{https://github.com/PurdueDigitalTwin/MACP}.
\end{abstract}

\section{Introduction}
\label{sec:intro}
Automated vehicles have made significant progress in recent years. However, their perception systems fall short in occlusions and long-range perception, hindering higher-level autonomy. Recently, cooperative perception systems through vehicle-to-vehicle (V2V) communications have emerged as a promising solution. This approach shifts the perception from a single-agent perspective into a joint task of the connected and automated vehicles (CAVs) system. It can address occlusion challenges and enhance the overall perception performance of individual vehicles~\cite{caillot_survey_2022}. Through collaborative information sharing, cooperative perception has the potential to revolutionize the automotive industry.

\begin{figure}[t]
    \centering
    \includegraphics[width=\linewidth]{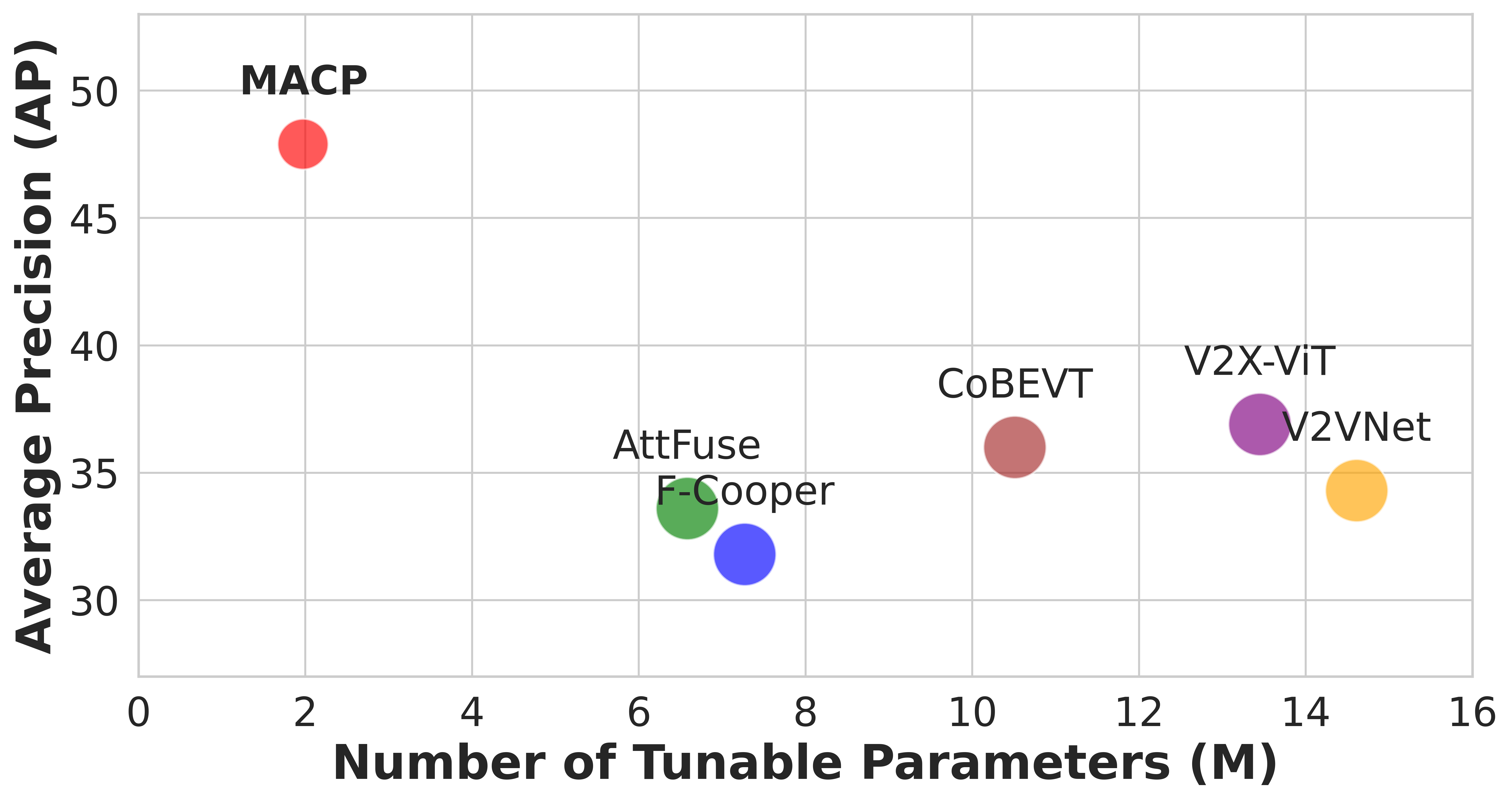}
    \caption{Performance comparison with state-of-the-art (SOTA) methods (F-Cooper~\cite{chen_f-cooper_2019}, V2VNet~\cite{wang_v2vnet_2020}, AttFuse~\cite{xu_opv2v_2022}, V2X-ViT~\cite{xu_v2x-vit_2022}, CoBEVT~\cite{xu_cobevt_2022}) on the V2V4Real Dataset~\cite{xu_v2v4real_2023}: Bubble size corresponds to the transmitted data size required by the algorithms. Our proposed MACP model outperforms the leading SOTA model by achieving a \textbf{30\%} improvement in Average Precision (AP) at Intersection over Union (IoU) = 70 while requiring only \textbf{15\%} of the number of tunable parameters and \textbf{65\%} of the volume of data transmission.}
    \label{fig:enter-label}
\end{figure}

The latest V2V cooperative perception studies explicitly aim to design dedicated perception models for the new setting. These sophisticated perception models have shown impressive performance improvements~\cite{xu_cobevt_2022,xu_v2x-vit_2022,qiao_adaptive_2023} over their single-agent counterparts. Nevertheless, developing and training these large cooperative perception models can be expensive. Moreover, extensive data labeling for training in the cooperative perception context can be time-consuming and costly, setting back the scalability of these approaches~\cite{zhao_epointda_2021,xu_v2v4real_2023}.

An intuitive way to reduce the cost is to take advantage of the generalization potential of existing single-agent perception models and adapt them to cooperative perception. Pre-trained single-agent models have demonstrated excellent transferability~\cite{cui_radar_2023}, raising the possibility of reusing their generalizable representations of the observations. Although promising, the feasibility of this adaptation process remains questionable, as inappropriate fine-tuning using downstream data could compromise the pre-trained model's performance. Therefore, our framework draws inspiration from the parameter-efficient fine-tuning (PEFT) strategy, initially explored in natural language processing~\cite{houlsby_parameter-efficient_2019}. The PEFT strategy aims to retain the pre-trained model's strength by only fine-tuning a small number of additional parameters while freezing the majority of the model. 

This paper introduces a novel framework termed ``efficient Model Adaption for Cooperative Perception," or MACP. 
Specifically, we bridge the gap between single-agent and cooperative perception by addressing domain shifts and communication bottlenecks. We design the Convolution Adapter (ConAda) for the feature encoder and communication channel and add Scale and Shift the Features (SSF) operations~\cite{lian_scaling_2022} in the prediction net to mitigate domain shifts. Our experiments show that our framework significantly outperforms previous state-of-the-art (SOTA), with substantially fewer tunable parameters. Moreover, the proposed MACP framework inherently supports compressed data transmission and can effectively utilize shared data. In summary, we make the following contributions:

\begin{itemize}
    \item We identify the gap between the efficacy of single-agent perception models and the requirements of cooperative perception.
    \item We propose a novel framework to empower adapting single-agent perception models to the cooperative perception, which is simple to implement and cost-effective to train.
    \item The proposed model significantly outperforms previous SOTA methods on simulation and real-world cooperative perception benchmarks. Especially, a $30\%$ improvement is achieved in Average Precision (AP) at Intersection over Union (IoU) = 70 on the V2V4Real dataset~\cite{xu_v2v4real_2023}, with only $15\%$ of the number of tunable parameters and $65\%$ of data transmission size.
\end{itemize}

\section{Related Work}
\label{sec:related}

\subsection{3D Object Detection}
\label{sec:3dod}
Accurate object perception is crucial for ensuring the safety of autonomous driving systems. SOTA 3D object detection models commonly use sparse convolutions to extract point cloud features~\cite{yan_second_2018}. These features are then combined with either anchor-based~\cite{yan_second_2018, deng_voxel_2021} or center-based~\cite{yin_center-based_2021, zhou_centerformer_2022} strategies for making predictions.

VoxelNet~\cite{zhou_voxelnet_2018} encodes voxel features using PointNet~\cite{qi_pointnet_2017}, and then applies a region proposal network and a prediction head. SECOND~\cite{yan_second_2018} improves performance by integrating efficient sparse convolutions with an anchor-based prediction head. CenterPoint, inspired by CenterNet~\cite{duan_centernet_2019}, converts the sparse output from a backbone network into a feature map and then predicts object center locations through heatmap generation. This center-based prediction strategy has been adopted by various models~\cite{bai_transfusion_2022,zhou_centerformer_2022, liu_bevfusion_2023} to enhance the performance of 3D detection frameworks. However, single-agent 3D object detection models suffer from occlusions and long-range prediction. We tackle these through V2V cooperative perception. 

\subsection{Cooperative Perception}
\label{sec:coperception}
The fundamental idea of cooperative perception in CAVs is to enhance their field of view by sharing observations from surrounding vehicles or roadside infrastructures~\cite{caillot_survey_2022}. There are three categories of cooperation approaches based on their data-sharing strategies: 
(1) \textbf{Early Fusion}~\cite{chen_cooper_2019}: CAVs transmit raw sensor data, and the ego vehicle makes predictions based on the aggregated raw data, which incurs the highest data transfer cost;
(2) \textbf{Late Fusion}~\cite{rockl_v2v_2008,rauch_car2x-based_2012,xu_pointfusion_2018,zhang_distributed_2021,yu_dair-v2x_2022}: Late fusion involves sharing and combining predictions (\textit{e.g.}, 3D bounding boxes) from CAVs, reducing the data transfer load. However, the performance is dependent on the prediction precision of other CAVs and is sensitive to spatial and temporal misalignment introduced by issues like localization error and transmission delay;
(3) \textbf{Mid Fusion}~\cite{marvasti_cooperative_2020,wang_v2vnet_2020,li_learning_2021,xu_cobevt_2022,xu_v2x-vit_2022,qiao_adaptive_2023,xiang_hm-vit_2023}: Mid-fusion involves collaborators extracting intermediate features, encoding them and then compressing and broadcasting them to other vehicles. This strategy offers higher flexibility and has the potential to balance performance, robustness, and communication costs, and is therefore gaining more attention. For instance, researchers have proposed vision transformers with attention modules specifically designed for cooperative perception in V2X-ViT~\cite{xu_v2x-vit_2022} and CoBEVT~\cite{xu_cobevt_2022}. Qiao et al. have also proposed adaptive feature fusion models with trainable neural networks in AdaFusion~\cite{qiao_adaptive_2023}. Note that all these models are specifically designed for the cooperative setting and are trained from scratch, which can be costly. This paper distinguishes itself by exploring efficient adaptations of pre-trained single-agent perception models to address these concerns.



\subsection{Parameter-Efficient Fine-Tuning}
\label{sec:peft}
Implementing a pre-trained model for a different task often requires a fine-tuning procedure based on the new dataset. Essentially, the goal is to preserve the knowledge parameterized by the pre-trained model and adjust it to fit the application context. Previous studies have approached this objective from different perspectives, including multi-task learning with fixed upstream layers~\cite{caruana_multitask_1997, zhang_robust_2013, zhang_facial_2014, bilen_integrated_2016}, sequential life-long learning by adding new parameters~\cite{rusu_progressive_2018}, and preserving knowledge from the old task~\cite{kirkpatrick_overcoming_2017,li_learning_2018,zhao_end--end_2020}. 

Recent studies have shown that fine-tuning can be achieved by updating or appending a relatively small number of parameters. Fewer tunable parameters generally are more energy efficient ~\cite{desislavov_trends_2023} and enable fast iterative prototyping and transferring. The Adapter~\cite{rebuffi_learning_2017, houlsby_parameter-efficient_2019, bapna_simple_2019,yang_aim_2023} and its applications are among the earliest parameter-efficient fine-tuning (PEFT) methods. Adapter~\cite{houlsby_parameter-efficient_2019} is inserted into transformer~\cite{vaswani_attention_2017} layers, and it first projects and transforms features into a low-dimensional space and then projects them back to the source domain. Subsequent studies have extended this idea by using low-rank updates~\cite{hu_lora_2022} or sparse parameter selection~\cite{sung_training_2021, guo_parameter-efficient_2021}. SSF~\cite{lian_scaling_2022} is another PEFT method designed to scale and shift the features extracted by a pre-trained model. The idea behind SSF is to address the distribution mismatch between the upstream and downstream tasks. These SOTA PEFT methods have demonstrated the ability to significantly reduce the training cost of models while matching the performance of fully fine-tuning all the parameters. However, adapting existing single-agent perception models for cooperative perception contexts remains underexplored, a gap that this paper aims to address.

\begin{figure*}[!t]
    \centering
    \includegraphics[width=0.85\linewidth]{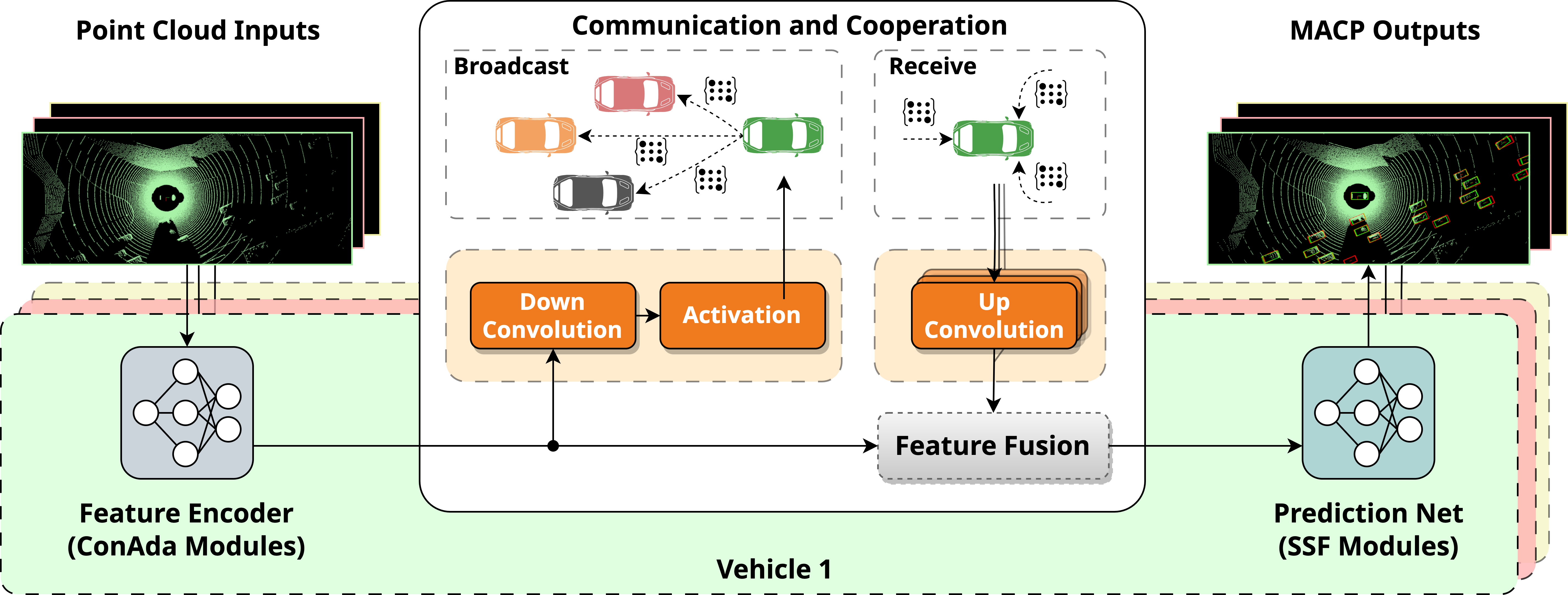}
    \caption{\textbf{Overview of the MACP Framework.} The down convolution and the activation layers compress the output from the feature encoder to enable efficient communication with surrounding vehicles. Meanwhile, each vehicle receives features from its surroundings and decompresses the information using the up convolution layer in a stack of ConAda modules. A feature fusion module fuses the decompressed data with the encoded feature from the upstream feature encoder and passes it to the downstream prediction network.}
    \vspace{-2ex}
    \label{fig:coperation}
\end{figure*}

\section{Methodology}
\label{sec:method}


\subsection{Problem Statement} 
\label{sec:problem-statement}

This paper focuses on cooperative 3D object detection through V2V communication. Given a traffic condition with $N$ agents, each agent $i$ has an observed point cloud $x_i\in\mathbb{R}^{N_i\times d}$ consisting of $N_i$ points in 3D Euclidean space. We represent each point by a $d$-dimensional vector with attributes such as coordinates and reflected intensity. We denote the set of all point clouds by $\mathbf{X}=\left\{x_i\right\}_{i=1,\ldots,N}$. The objective is to solve for an optimal model $f^*$ capable of detecting and delineating bounding boxes about surrounding objects and assigning appropriate labels. To simplify our notations, we represent each bounding box and its class label by a $d^\prime$-dimensional vector $y_j\in\mathbb{R}^{d^\prime}$. Without losing generality, an object detection model $f$ is a mapping from point cloud space to the joint space of bounding boxes and their labels $f:\mathcal{X}\rightarrow\mathcal{Y}$ and the trained model ideally describes the probability of observing bounding box set $\mathbf{y}$ conditioned on observed point cloud set $\mathbf{x}$, given by
\begin{equation}
    p(\mathbf{y}|\mathbf{x};f)=\frac{p(\mathbf{x},\mathbf{y})}{p(\mathbf{x})}.
\end{equation}

To adapt a pre-trained single-agent perception model for the cooperative perception setting, we need to deal with distribution shifts in observed point clouds. Specifically, if we denote the marginal probability of observing a point cloud set in single-agent perception by $p_\mathcal{S}(\mathbf{x})$ and the probability of observing the exact set in a cooperative perception by $p_\mathcal{C}(\mathbf{x})$, the two probabilities can be different due to additional point clouds shared by V2V communication, that is, $p_\mathcal{S}(\mathbf{x})\neq p_\mathcal{C}(\mathbf{x})$. The joint distribution of point clouds and bounding boxes given by the pre-trained model deviates from the ground-truth joint distribution under the cooperative setting:
\begin{equation}
    \hat{p}_\mathcal{C}(\mathbf{x},\mathbf{y};f)=\frac{p_\mathcal{S}(\mathbf{x},\mathbf{y})}{p_\mathcal{S}(\mathbf{x})}\cdot p_\mathcal{C}(\mathbf{x})\neq p_\mathcal{C}(\mathbf{x},\mathbf{y}).
\end{equation}
Existing literature~\cite{sugiyama_covariate_2007,koh_wilds_2021} refers to this phenomenon as domain shifts and has shown that it directly leads to performance degradation. In this paper, we propose to introduce a collection of light-weight fine-tuning module $g$ and transform $p(\mathbf{y}|\mathbf{x};f)$ such that $p(\mathbf{y}|\mathbf{x};f\cdot g)=g\left[\frac{p_\mathcal{S}(\mathbf{x},\mathbf{y})}{p_\mathcal{S}(\mathbf{x})}\right]$ and
\begin{equation}
    g^*=\underset{g\in\mathcal{G}}{\text{argmin}}\mathcal{L}\left(p_\mathcal{C}(\mathbf{x},\mathbf{y}), \hat{p}_\mathcal{C}(\mathbf{x},\mathbf{y};f\cdot g)\right),
\end{equation}
where $\mathcal{L}$ is a loss function measuring the distance between two distributions.


Meanwhile, cooperative perception introduces new challenges in terms of computation and communication. The model must handle compressed data with minimal performance loss to ensure stable data transmission and responsive decision-making in high-traffic environments. This paper presents the MACP framework with specific design and implementation of PEFT modules, following guidelines to account for domain shifts and communication bottlenecks while maximizing performance with minimal trainable parameters. The following sections introduce two PEFT modules, ConAda and SSF Operators, followed by elaborations on how we address the guidelines in their designs and implementations.





\subsection{MACP Overview}
\label{sec:overview}
As illustrated in \cref{fig:coperation}, the proposed MACP is a decentralized framework where each vehicle encodes point cloud features locally using a Feature Encoder network with ConAda modules. After local feature encoding, the vehicles communicate with each other via a compression-decompression channel driven by another ConAda module. We provide further information about the ConAda module in \cref{sec:conada}. The collected features are fused with the feature from the local Feature Encoder, and then passed through a Prediction Net with SSF modules (refer to \cref{sec:ssf-operator}) to generate bounding box predictions. Note that in our settings, all vehicle models share the same parameters.

\subsection{Convolution Adapter}
\label{sec:conada}
\begin{figure*}[!t]
    \centering
    \includegraphics[width=0.85\textwidth]{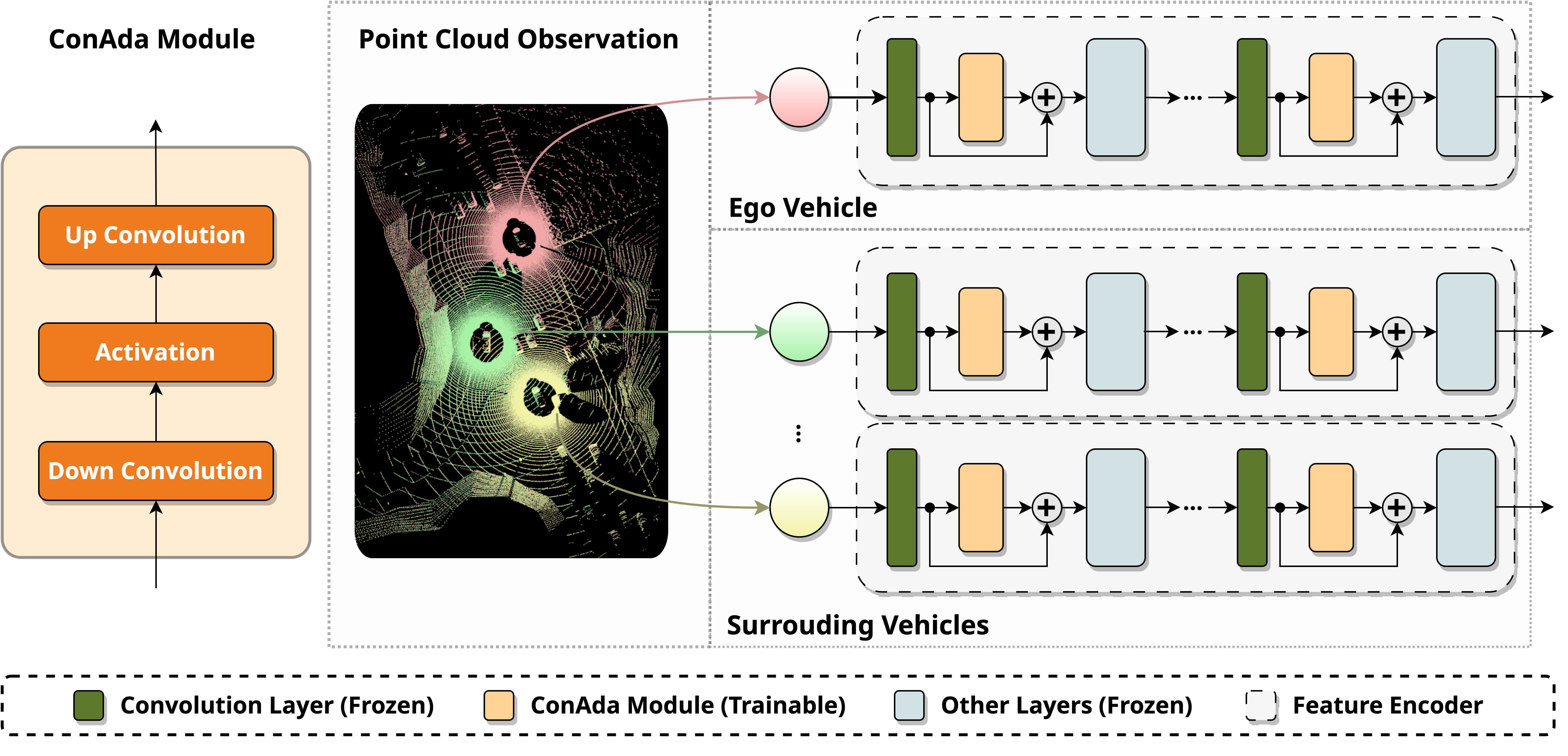}
    \caption{\textbf{Illustration of the Distributed Feature Encoder with ConAda module.} Ego and surrounding vehicles each encode their observations with a feature encoder consisting of a cascade of blocks. All feature encoders share the same parameters. In each block, a ConAda module (yellow) processes the output feature map from the pre-trained convolution layer, adds it back to the convolution output through residual connection, and passes it to the consecutive other layers.}
    \label{fig:v2v_conada}
\end{figure*}

Mainstream 3D object detection models commonly rely on convolutional layers to capture local feature correlations effectively. To adapt the pre-trained single-agent parameters to the cooperative setting, we propose the Convolution Adapter (ConAda) module. As shown in~\cref{fig:v2v_conada}, ConAda consists of a down convolution layer that first projects the source feature map to a lower dimension. Following this projection, a non-linear activation and a subsequent up convolution layer transform and remap the lower-dimension feature back to the source space. Given an input feature map $\mathbf{I}$, the output feature map of ConAda is calculated as follows
\begin{equation}
\label{eq:conada}
\mathbf{O}=\text{Conv}(\text{Activation}(\text{Conv}(\mathbf{I}, \mathbf{K}^\text{down})), \mathbf{K}^\text{up}),
\end{equation}
where $\mathbf{K}^\text{down}\in\mathbb{R}^{D\times D'}$ and $\mathbf{K}^\text{up}\in\mathbb{R}^{D'\times D}$ are the down-projection and up-projection kernels respectively, and their dimensions satisfy $D'<D$. $\text{Activation}(\cdot)$ is the non-linear activation function, and $\text{Conv}(\cdot, \cdot)$ is the convolution operation with both kernel size and strides equal to 1.

Note that while visual data like images are inherently dense, 3D point clouds derived from LiDAR are intrinsically sparse. Applying dense convolutional operations on such data is computationally inefficient~\cite{graham_3d_2018}. As a result, 3D object detectors commonly use sparse convolutions to extract point cloud features (see \cref{sec:3dod} for detailed descriptions). To this end, our proposed ConAda module implementation inherently rests on sparse convolution operations. Specifically, we use the Submanifold sparse convolution~\cite{graham_submanifold_2017} within the ConAda module. Specifically, we consider a sparse voxelized input $X^\text{input} = \{ (v_1, f_1), (v_2, f_2), \dots, (v_N, f_N) \}$, where $v_i$ is the voxel's coordinates and $f_i\in C$ is the associated feature vector. Given the kernel $\mathbf{K}\in\mathbb{R}^{C\times C^\prime}$, for an \textit{occupied} voxel $v_i$ with feature $f_i$, the output feature vector after the convolution is
\begin{equation}
    {X_{i,c'}^\text{output}} = \sum_{k=1}^{C} \mathbf{K}_{k,c'} \times f_{i,k},
\end{equation}
where ${X_{i,c'}^\text{output}}$ is the $c'$-th element of the output feature vector for voxel $v_i$.

ConAda modules are key components for the Feature Encoder. A Feature Encoder network is a cascade of convolution blocks where the output from the convolution layer passes a ConAda module and adds back to itself with a residual connection~\cite{he_deep_2016}. We only train the ConAda parameters during training and freeze the pre-trained parameters in the convolution layer and other layers following the ConAda module.

Meanwhile, ConAda also acts as the communication channel between vehicles. During communications, the down convolution and the activation layer in the ConAda module help compress and encrypt the encoded feature for broadcasting, while the up convolution layer serves to decompress the received signal for feature fusion.

\subsection{SSF Operator for Fused Feature}
\label{sec:ssf-operator}
The output from the point encoder module is a latent feature map representing a mixture of point clouds observed by ego and surrounding vehicles.
We implement the SSF~\cite{lian_scaling_2022} operator in the consecutive neural network layers to account for the domain shift. Suppose the output feature map from the convolution layer is given by $X^\text{output}\in\mathbb{R}^{H^\prime\times W^\prime\times C^\prime}$, we update the feature map by using a scaling factor $\gamma\in\mathbb{R}^{C^\prime}$ and a shifting factor $\beta\in\mathbb{R}^{C^\prime}$, given as
\begin{equation}
    \label{eq:ssf}
    X^\text{output}_{i, j} = \gamma\ \odot\ X^\text{output}_{i, j} + \beta,
\end{equation}
where $\odot$ is the Hadamard product. The SSF operator only scales and shifts the feature without altering their positional identities.

\begin{figure}[!t]
    \centering
    \includegraphics[width=\linewidth]{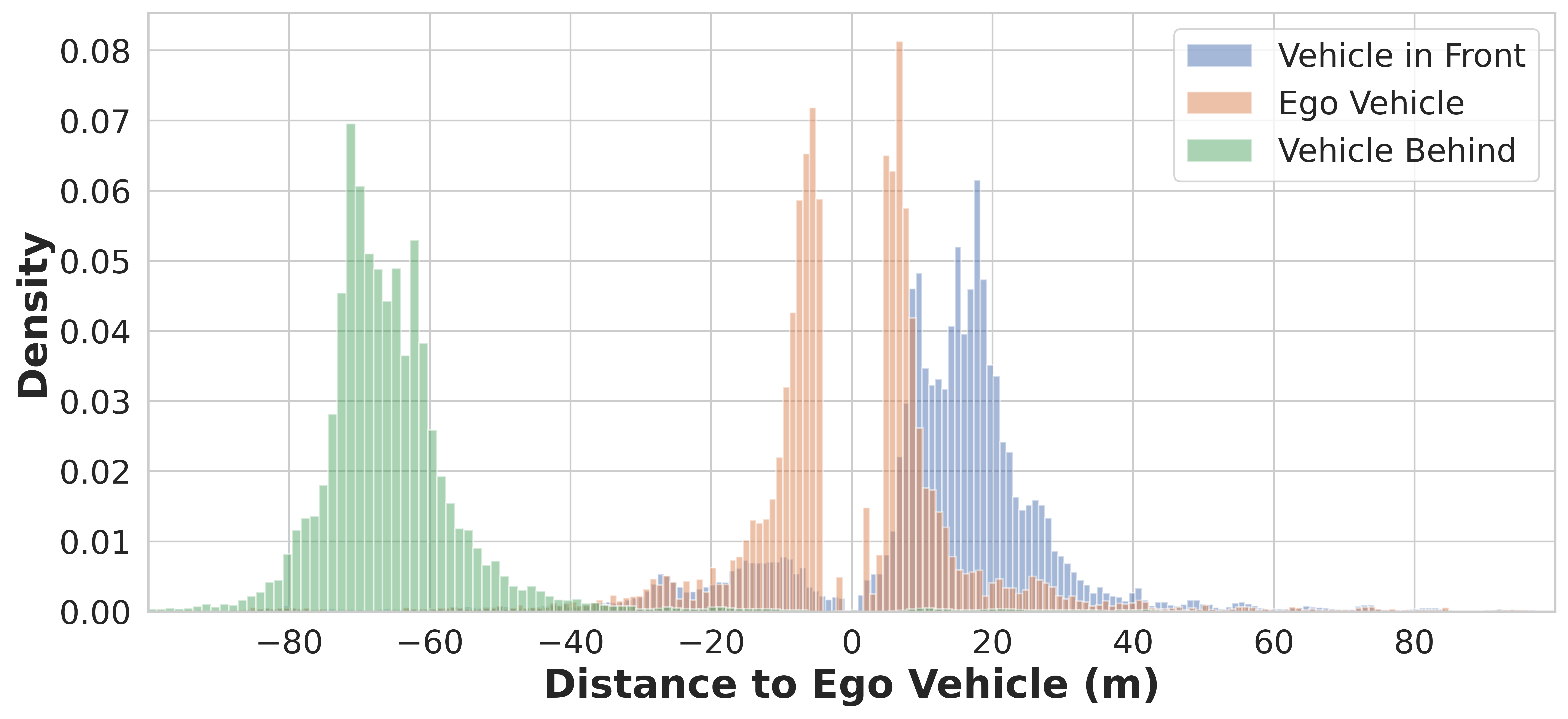}
    \caption{\textbf{Example of spatial distribution shifts of point clouds.} On the x-axis, the distance's sign is determined by the angle between the ego vehicle's direction and the line from it to the point. The result indicates a significant difference in the data distribution from the surrounding vehicles from that of the ego vehicle.}
    \vspace{-2ex}
    \label{fig:dist-shift}
\end{figure}

\subsection{Insights}
\label{sec:insights}
As mentioned in \cref{sec:problem-statement}, our MACP framework and the two PEFT modules account for domain shifts while considering computation and communication constraints. We focus on covariate shifts of point cloud features for domain shifts and identify two primary sources: spatial distribution shifts and feature space shifts.

Spatial distribution originates from point clouds collected from different vehicles concentrating around their respective locations and differing in scales due to varying sensors, as shown in \cref{fig:dist-shift}. A simple concatenation of these point clouds leads to a shift in the spatial distribution of the data from a bi-modal to a multi-modal mixture distribution. However, since the pre-trained model was trained on the bi-modal or even an uni-modal point cloud distribution, the parameters may not recognize the importance of out-of-distribution point cloud features. Our MACP addresses this issue through the distributed framework. In this way, we enforce each vehicle to encode point cloud features from their local perspectives, which aligns more with the single-agent perception cases and can help eliminate impacts from the out-of-distribution problem.

Feature space shift is manifested as each latent feature map element may contain more or less information compared to the single-agent perception. This requires extra operators to scale and shift the feature maps to align with the latent space from single-agent perception before the direct use of pre-trained parameters. ConAda modules in the MACP framework approach this by projecting and applying non-linear transformation in a low-dimensional space, while the SSF modules directly apply scales and shifts to the input feature maps.

Finally, the ConAda-based communication channel mitigates the communication bottlenecks by allowing flexible compression of the signal to transmit. We will investigate and show in our experiments how this compression-decompression process affects the performance.

\section{Experiments and Results}
\label{sec:expriment}
\begin{table*}[!t]
    \centering
    \small
    \resizebox{0.8\linewidth}{!}{
    \begin{tabular}{l|cc|cccc|c}
        \hline
        \multirow{2}{4em}{Method} & \multicolumn{2}{c|}{Param (M)} & \multicolumn{4}{c|}{AP@IoU=50/70 $(\uparrow)$} & AM $(\downarrow)$\\
        &Total&Trainable & Overall & 0-30m & 30-50m & 50-100m & (MB)\\
        \hline
        No Fusion & 6.58 & 6.58 & 39.8/22.0 & 69.2/42.6 & 29.3/14.4 & 4.8/1.6 & 0 \\
        Late Fusion & 6.58 & 6.58 & 55.0/26.7 & 73.5/36.8 & 43.7/22.2 & 36.2/17.3 & 0.003 \\
        Early Fusion & 6.58 & 6.58 & 59.7/32.1 & 76.1/46.3 & 42.5/20.8 & \underline{47.6}/\underline{21.1} & 0.96 \\
        \hline
        F-Cooper~\cite{chen_f-cooper_2019}  & 7.27 & 7.27  & 60.7/31.8 & 80.8/46.9 & 45.6/23.6 & 32.8/13.4 & 0.20\\
        V2VNet~\cite{wang_v2vnet_2020}  & 14.61 & 14.61  & 64.5/34.3 & 80.6/51.4 & 52.6/26.6 & 42.6/14.6 & 0.20\\
        AttFuse~\cite{xu_opv2v_2022}  & 6.58 & 6.58  & 64.7/33.6 & 79.8/44.1 & \underline{53.1}/\underline{29.3} & 43.6/19.3 & 0.20\\
        V2X-ViT~\cite{xu_v2x-vit_2022}  & 13.45 & 13.45  & 64.9/\underline{36.9} & 82.0/\underline{55.3} & 51.7/26.6 & 43.2/16.2 & 0.20\\
        CoBEVT~\cite{xu_cobevt_2022}  & 10.51 & 10.51 & \underline{66.5}/36.0 & \underline{82.3}/51.1 & 52.1/28.2 & \textbf{49.1}/19.5 & 0.20\\
        MACP \textbf{(Ours)} & 8.94 & 1.97 & \textbf{67.6}/\textbf{47.9} & \textbf{83.7}/\textbf{62.1} & \textbf{58.4}/\textbf{38.5} & 34.6/\textbf{23.1} & 0.13\\
        \hline
    \end{tabular}
    }
    \vspace{-1ex}
    \caption{\textbf{Performance Comparison on the V2V4Real Dataset}. The best-performing method is highlighted in \textbf{bold}, while the second-best method is indicated by an \underline{underline}. $\downarrow$: Lower values are better. $\uparrow$: Higher values are better.}
    \label{tab:sota-v2v4real}
\end{table*}

\begin{table}[!t]
    \centering
    \small
    \resizebox{\linewidth}{!}{
    \begin{tabular}{l|cc|cc}
        \hline
        \multirow{2}{4em}{Method} & \multicolumn{2}{c|}{Param (M)} & \multicolumn{2}{c}{AP@IoU=50/70 $(\uparrow)$}\\
        & Total & Trainable & Default Towns & Culver City\\
        \hline
        No Fusion & 6.58 & 6.58 & 67.9/60.2 & 55.7/47.1 \\
        Late Fusion & 6.58 & 6.58 & 85.8/78.1 & 79.9/66.8 \\
        Early Fusion & 6.58 & 6.58 & 89.1/80.0 & 82.9/69.6 \\
        \hline
        F-Cooper~\cite{chen_f-cooper_2019} & 7.27 & 7.27 & 88.7/79.0 & 84.6/72.8\\
        V2VNet~\cite{wang_v2vnet_2020} & 14.61 & 14.61 & 89.7/82.2 & 86.0/73.4\\
        AttFuse~\cite{xu_opv2v_2022} & 6.58 & 6.58 & 90.8/81.5 & 85.4/73.5\\
        V2X-ViT~\cite{xu_v2x-vit_2022} & 13.45 & 13.45 & 89.1/82.6 & 87.3/73.7\\
        CoBEVT~\cite{xu_cobevt_2022} & 10.51 & 10.51 & 91.4/\underline{86.1} &85.9/77.2\\
        AdaFusion~\cite{qiao_adaptive_2023} & 7.27 & 7.27 & \underline{91.6}/85.6 & \underline{88.1}/\underline{79.0}\\
        MACP \textbf{(Ours)} & 8.98 & 2.00 & \textbf{93.7}/\textbf{90.3} & \textbf{91.4}/\textbf{80.7}\\
        \hline
    \end{tabular}
    }
    \vspace{-1ex}
    \caption{\textbf{Performance Comparison on the OPV2V Dataset.}}
    \vspace{-2ex}
    \label{tab:sota-opv2v}
\end{table}

\subsection{Experimental Settings}
\label{sec:exp-dataset}
\paragraph{Datasets}
We conducted comprehensive experiments on two widely used cooperative perception benchmarks: the V2V4Real~\cite{xu_v2v4real_2023} and the OPV2V~\cite{xu_opv2v_2022,xu_opencda_2023} datasets. Both datasets support the 3D object detection task where different types of vehicles are considered as the same category, and the detection targets are vehicles. Each CAV comes with its individual LiDAR point cloud and corresponding ground-truth 3D bounding boxes. 

\textit{OPV2V} is a simulation-based dataset that includes two subsets: Default Towns (DT) and Culver City (CC). The DT subset contains data from eight default towns provided by CARLA~\cite{dosovitskiy_carla_2017}. Each frame contains an average of about 3 CAVs, with a minimum of 2 and a maximum of 7. It has an official split for training, validation, and testing with 6.7K, 2K, and 2.7K frames, respectively. The CC scenes form a separate test set of 550 frames, evaluating the model's generalization capability in unseen scenes. \textit{V2V4Real} is a real-world dataset collected by two vehicles driving simultaneously in Columbus, Ohio, USA. It is officially split into the train, validation, and test sets with 14K, 2K, and 4K frames, respectively.

\paragraph{Evaluation}
For fair comparisons, we use the same settings as in previous studies. Specifically, for the OPV2V dataset, we set the evaluation range in the $x$ and $y$ directions to (-140m, 140m) and (-40m, 40m), respectively, relative to the ego vehicle. For the V2V4Real dataset, the evaluation range in the $x$ and $y$ directions is set to (-100m, 100m) and (-40m, 40m), respectively, with reference to the ego vehicle. We evaluate the detection performance using Average Precision (AP) at Intersection-over-Union (IoU) 0.5 and 0.7 as the metric. Following~\cite{xu_v2v4real_2023}, we use the Average MegaByte (AM) metric to quantify the volume of transmitted data of algorithms on the V2V4Real dataset. All models are evaluated under the \textit{Sync} setting, where data transmission is considered instantaneous~\cite{yu_dair-v2x_2022,xu_v2v4real_2023}. 

\paragraph{Implementation}
We use the \textit{LiDAR-only BEVFusion}~\cite{liu_bevfusion_2023} as the single-agent perception model, 
which was pre-trained on the nuScenes~\cite{caesar_nuscenes_2020} dataset. 
The ConAda modules utilize GELU~\cite{hendrycks_gaussian_2016} as the activation function. 
Optimization is performed using AdamW~\cite{loshchilov_decoupled_2019} with a weight decay of $10^{-2}$. 
For more details, please refer to the Appendix.

\begin{table*}[!t]
    \begin{center}
        \resizebox{0.75\linewidth}{!}{
            \begin{tabular}{c|cc|cccc}
                \hline
                \multirow{2}{3em}{Method} & \multicolumn{2}{c|}{Param (M)} &\multicolumn{4}{c}{AP@IoU=50/70 $(\uparrow)$}\\
                &Total & Trainable  & Overall & 0-30m & 30-50m & 50-100m \\
                \hline
                Full Fine-Tune & 8.92 & 8.92 & 68.3/\textbf{51.9} & \textbf{85.0}/\textbf{65.7} & \textbf{59.1}/\textbf{41.2} & 33.0/\textbf{26.9}\\
                Train Fusion \& Head Only& 8.92 & 1.94 & 65.5/42.1 & 83.1/57.4 & 51.9/30.0 & 33.5/19.9\\
                \hline
                Adapter Only~\cite{houlsby_parameter-efficient_2019} & 9.17 & 2.19 & 63.8/37.5 & 81.0/50.8 & 49.1/26.5 & 33.7/17.6\\
                SSF Only~\cite{lian_scaling_2022} & 8.92 & 1.95 & 64.7/44.2 & 82.7/60.2 & 50.8/30.9 & 32.6/21.0\\
                ConAda Only & 8.97 & 2.00 & 67.5/49.0 & \underline{84.1}/62.2 & 57.0/38.6 & \underline{34.1}/25.6\\
                MACP & 8.98 & 2.00 & \textbf{69.4}/\underline{49.6} & 83.2/\underline{63.1} & \underline{58.6}/\underline{40.0} & \textbf{38.5}/\underline{26.5}\\
                \hline
            \end{tabular}
        }
    \end{center}
    \vspace{-2ex}
    \caption{\textbf{Effectiveness of proposed components.}} 
    \label{tab:peft}
\end{table*}

\subsection{Comparison with the State of the art}
In this section, we compare the proposed MACP method with previous studies, including baseline models such as no fusion, late fusion, early fusion, and several SOTA mid-fusion models (see \cref{sec:coperception}). 

\subsubsection{Results on V2V4Real}
\label{sec:sota-v2v4real}
\cref{tab:sota-v2v4real} presents performance comparisons on V2V4Real. Our method achieves superior performance by tuning only 1.97M parameters, which is much less than previous models, thanks to the knowledge successfully transferred from the pre-trained single-agent model. The data transmission size is 0.13 AM, which is 35\% lower compared to other mid-fusion methods and 87\% lower compared to the early fusion baseline. This size is implemented with a compression factor of 256 (see \cref{eq:compression-factor}). Despite the high compression factor, our method still achieves an overall AP score of 47.9 at IoU=70, which is a 30\% improvement over the runner-up method V2X-ViT~\cite{xu_v2x-vit_2022}. These results suggest that efficient finetuning can successfully handle cooperation.

In addition, we observed that MACP performs exceptionally well in high-precision predictions (IoU=70). It outperforms other methods across all object distance ranges: 0-30m, 30-50m, and 50-100m, with margins of $12\%$, $31\%$, and $9\%$ over other leading SOTA methods. When considering predictions with lower precision (IoU=50), our method continues to outperform its counterparts in the 0-30m and 30-50m ranges and the overall average precision (AP) metric. However, it is worth noting that our model's AP for objects in the 50-100m range is $30\%$ lower compared to the best-performing method CoBEVT~\cite{xu_cobevt_2022}. This discrepancy is due to biases in the pre-trained model and will be analyzed in detail in \cref{sec:module-effect}.

\subsubsection{Results on OPV2V}
\label{sec:sota-opv2v}
As shown in \cref{tab:sota-opv2v}, our method with 2M tunable parameters outperforms all prior methods. Specifically, in the Default Towns where the models are trained and tested, our approach shows a performance increase of 2.3\% and 4.8\% in terms of AP at IoU=50/70, respectively, compared to other methods. It is worth noting that the pre-trained model is trained on real-world data while the new domain is of simulated data. This suggests that our MACP method is able to bridge the gap not only between the single-agent and cooperative settings but also between real-world and simulation domains. Furthermore, in the Culver City scenes containing previously unseen environments during training, our model continues to outperform the leading SOTA method AdaFusion~\cite{qiao_adaptive_2023}, demonstrating improvements of 3.7\% and 2.1\% in terms of AP at IoU=50/70, respectively. These results show our model's great generalization ability.

\begin{figure}[!t]
    \centering
    \includegraphics[width=0.75\linewidth]{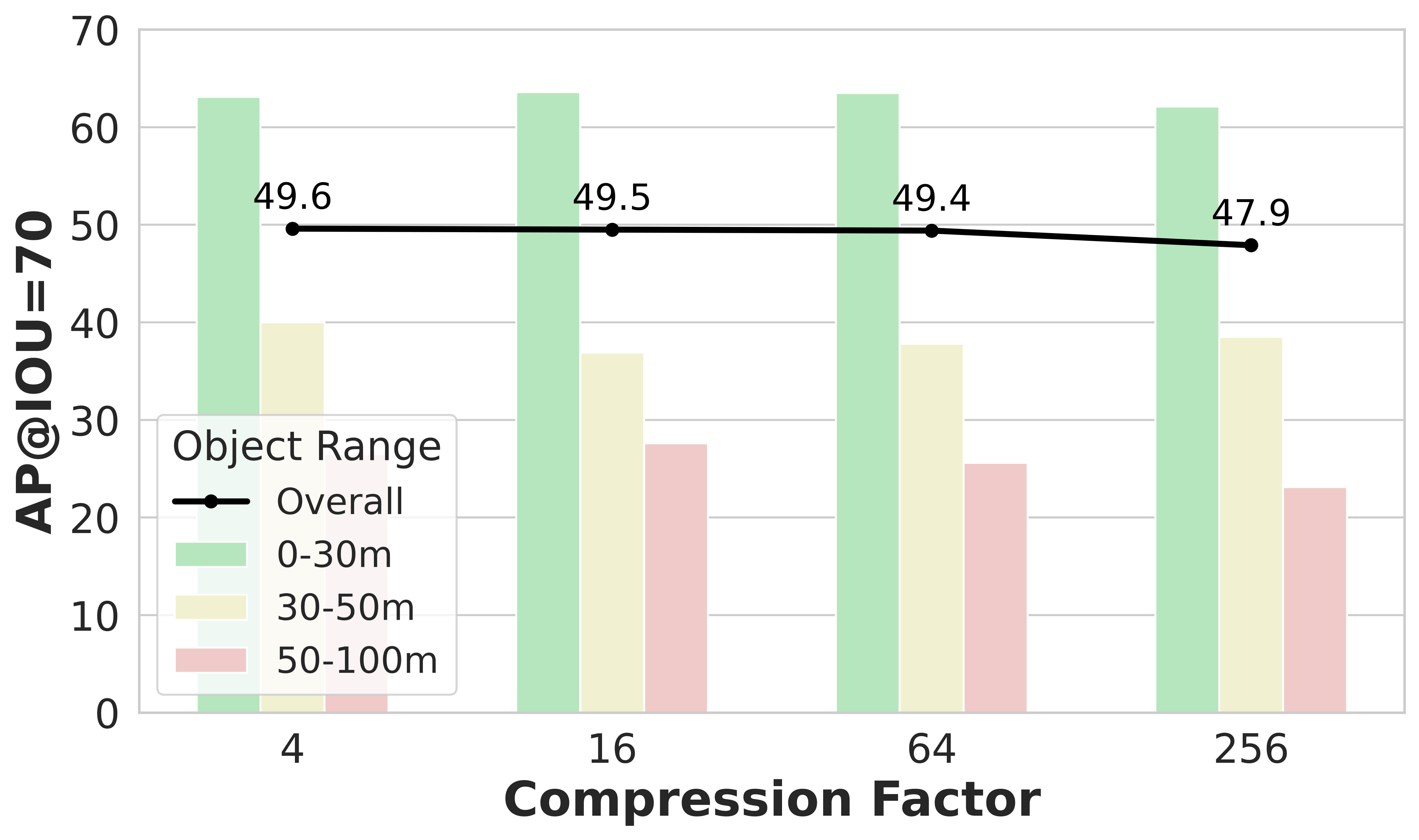}
    \caption{\textbf{Effect of compression factors.} 
    The visualization shows how the ConAda compression impacts the overall AP (black line) and AP of bounding boxes in different ranges (color bars).
    }
    \vspace{-1ex}
    \label{fig:compression}
\end{figure}

\begin{figure}[!t]
    \centering
    \includegraphics[width=0.9\linewidth]{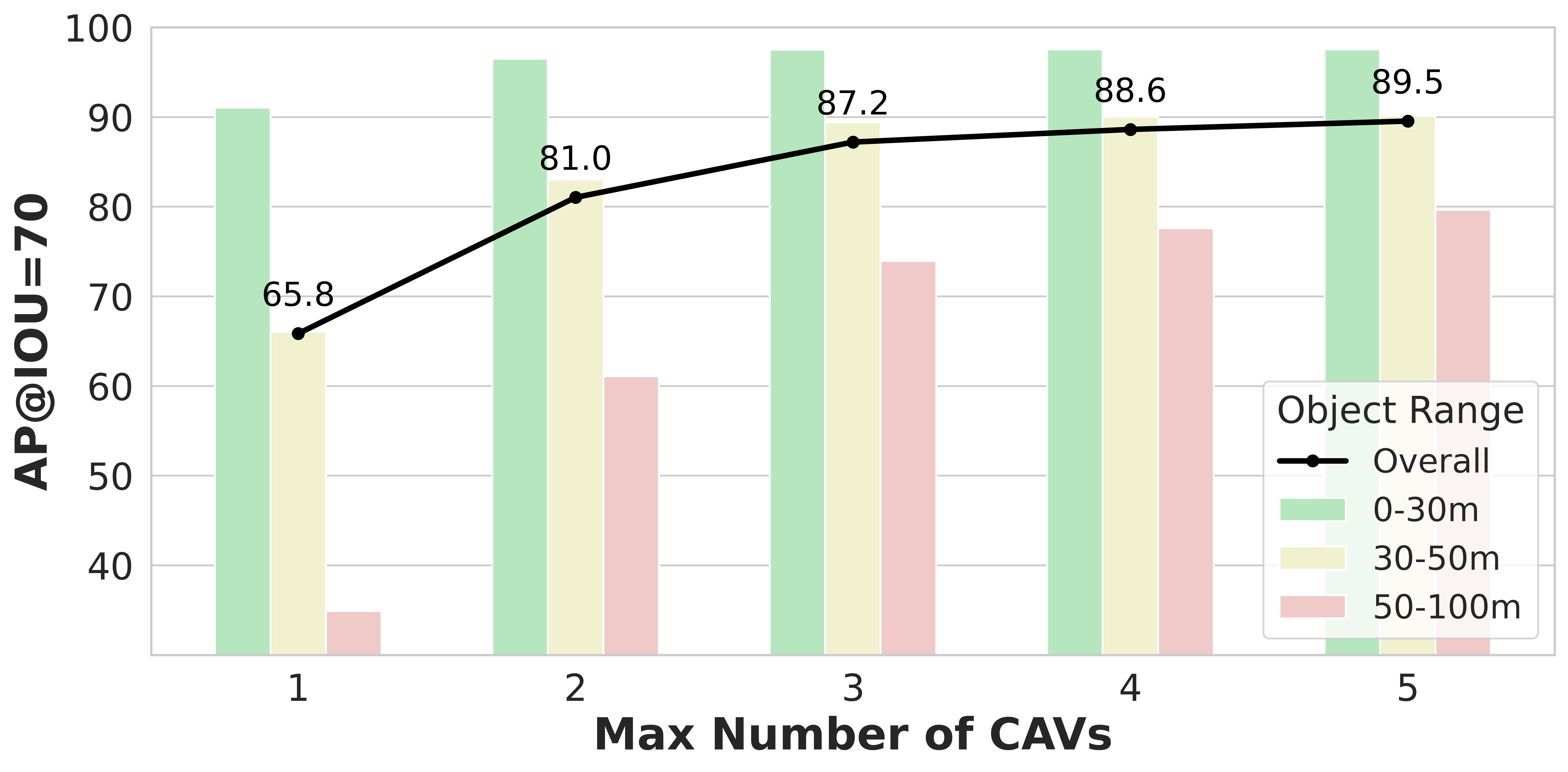}
    \caption{\textbf{Effect of max number of CAVs.}
    The upward trend of AP shows that the additional information from cooperative perception is being used effectively.
    \vspace{-2ex}
    } 
    \label{fig:num-cav}
\end{figure}

\begin{figure*}[!ht]
\centering
\begin{subfigure}{0.9\textwidth}
\centering
    \begin{subfigure}{0.49\linewidth}
        \rotatebox[origin=c]{90}{\includegraphics[trim={17cm 15cm 14.5cm 20cm},clip,height=\textwidth]{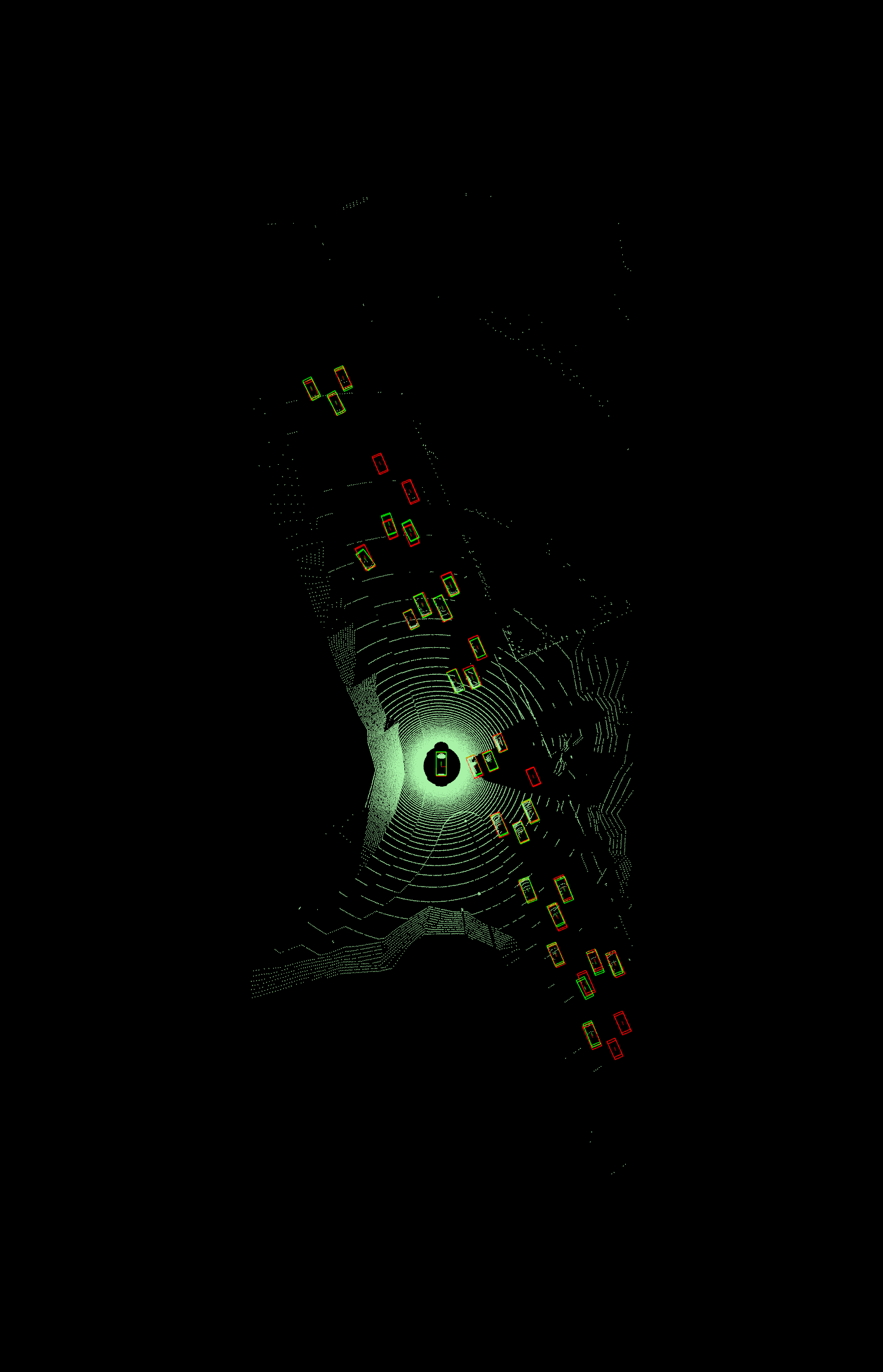}}
        \vspace{-2.5cm}
    \end{subfigure}
    \hfill
    \begin{subfigure}{0.49\linewidth}
        \rotatebox[origin=c]{90}{\includegraphics[trim={16cm 13cm 15.5cm 22cm},clip,height=\textwidth]{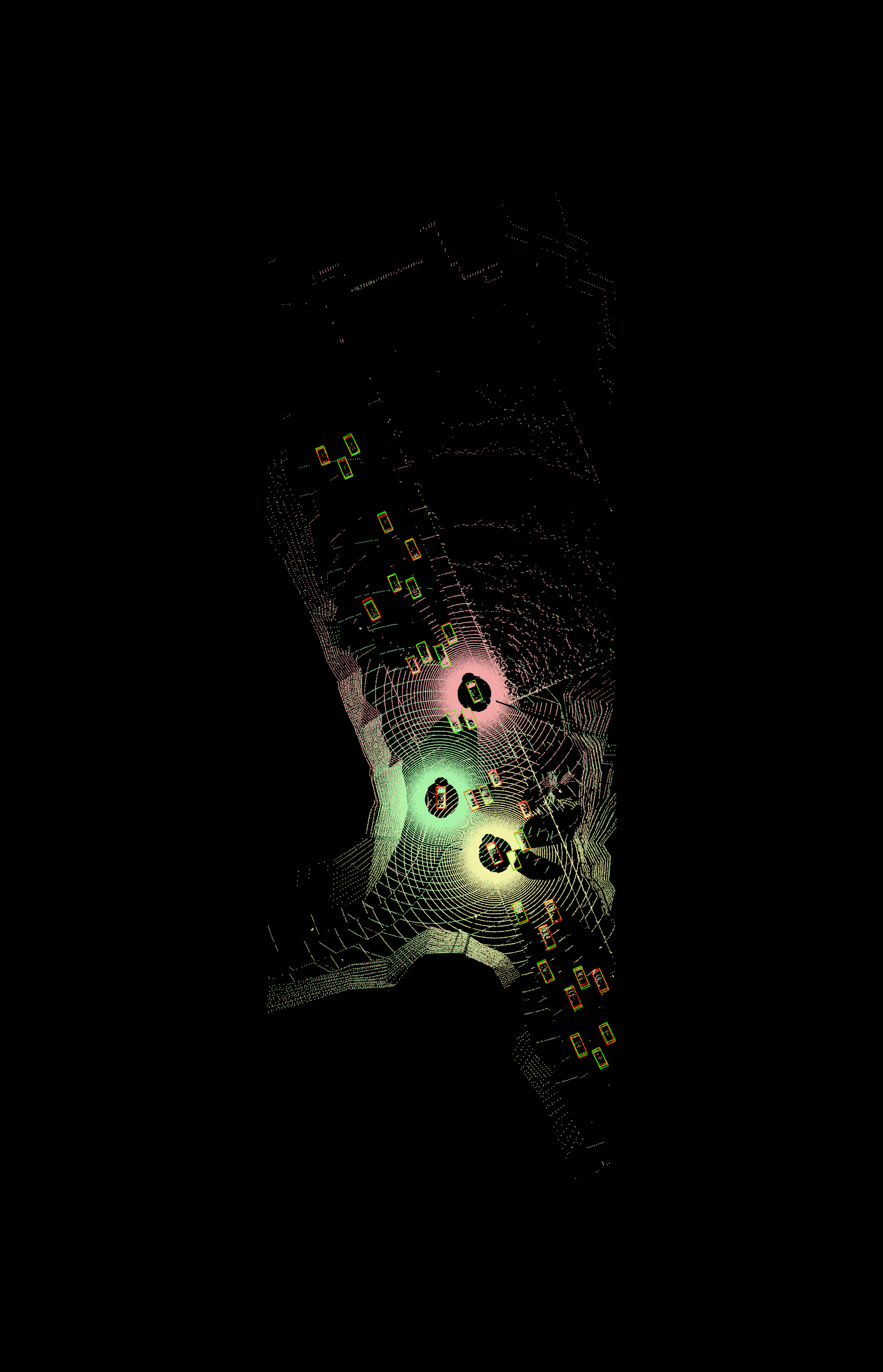}}
        \vspace{-2.5cm}
    \end{subfigure}
    \begin{subfigure}{0.49\linewidth}
        \rotatebox[origin=c]{90}{\includegraphics[trim={9cm 15cm 9cm 15cm},clip,height=\textwidth]{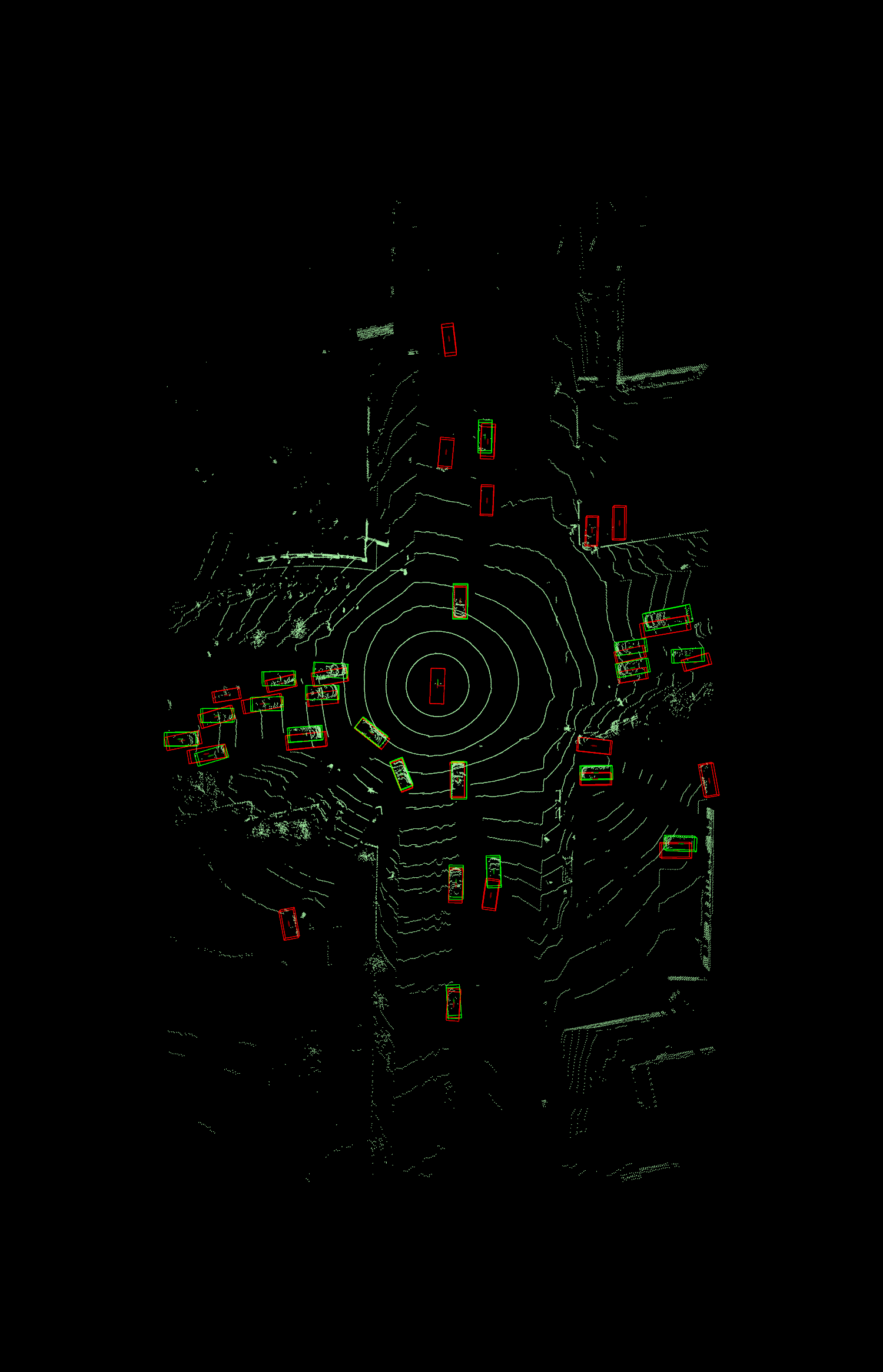}}
        \vspace{-1.5cm}
        \caption{BEVFusion~\cite{liu_bevfusion_2023} (Single-Agent Perception)}
    \end{subfigure}
    \hfill
    \begin{subfigure}{0.49\linewidth}
        \rotatebox[origin=c]{90}{\includegraphics[trim={9cm 15cm 9cm 15cm},clip,height=\textwidth]{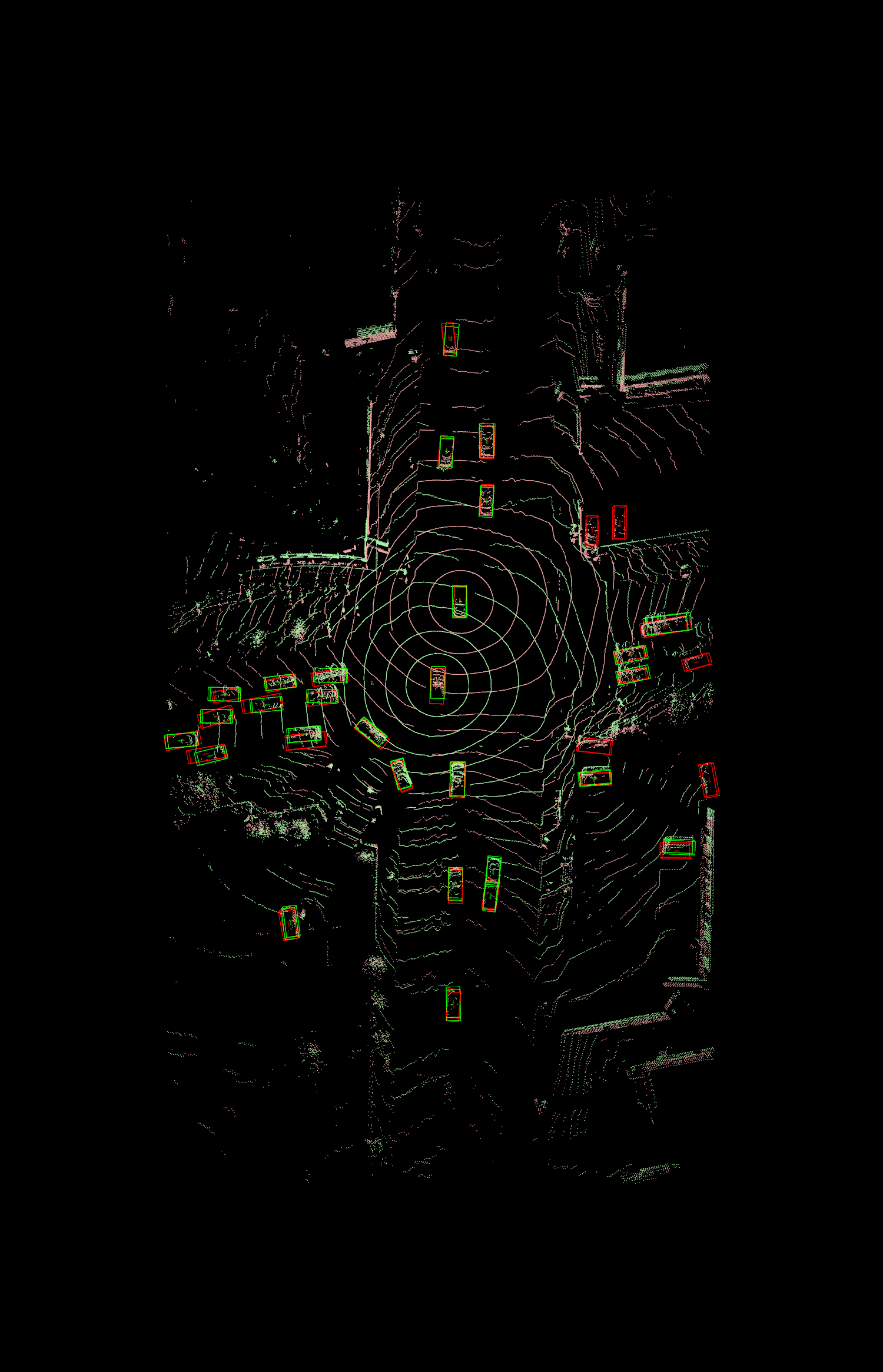}}
        \vspace{-1.5cm}
        \caption{MACP (Cooperative Perception)}
    \end{subfigure}
\end{subfigure}
\caption{Comparison between BEVFusion (Single-Agent Perception) and our MACP (Cooperative Perception) on the OPV2V (Top) and V2V4Real (Bottom) datasets. The point cloud of the ego vehicle is shown in \textcolor{YellowGreen}{light green}, while the point clouds of other vehicles are represented in different colors. The 3D bounding boxes in \textcolor{Red}{red} and \textcolor{green}{green} represent the \textcolor{Red}{ground-truth} and \textcolor{green}{predicted} objects, respectively. Our MACP outperforms BEVFusion in detecting occluded or distant objects.}
\vspace{-2ex}
\label{fig:vis}
\end{figure*}

\subsection{Module Effectiveness}
\label{sec:module-effect}
Since maximizing performance with minimal trainable parameters is one of our primary goals, this experiment aims to understand each proposed module's role in the overall performance and compare their parameter efficiency. We set up two baseline models: a fully fine-tuned BEVFusion model and another that only trains the fusion and bounding-box prediction heads. The second model is essentially the proposed framework but without any of the proposed PEFT modules. We compare the proposed MACP framework with three of its variants. The Adapter Only variant does not use the ConAda module in the feature encoder and replaces the SSF module in the Prediction Net with Adapter~\cite{houlsby_parameter-efficient_2019} modules. The SSF Only variant removes the ConAda modules from the feature encoder, and the ConAda Only variant removes the SSF modules from the Prediction Net of the proposed framework, respectively. All models are set with a compression factor of 4. \cref{tab:peft} details the total and tunable parameters and the overall and range-specific bounding box prediction accuracy.

We observe that the proposed MACP achieves comparable performance to the fully fine-tuned BEVFusion model, but with only around one-fifth of its total number of trainable parameters, indicating promising parameter efficiency. The ConAda Only variant outperforms the other two, suggesting that our original ConAda structure proposal plays a more significant role in the final performance. Ultimately, MACP benefits from both ConAda and SSF modules to achieve the best results.

Furthermore, it is worth noting that the fully fine-tuned model performs poorly when predicting distant objects (i.e., those in the range of 50-100 meters). This confirms that the poor long-distance prediction precision is due to the pre-trained parameters mentioned in \cref{sec:sota-opv2v}. In fact, our proposed MACP reduces the negative impact of the pre-trained bias and outperforms the fully fine-tuned model in long-distance predictions.

\subsection{Compression Robustness}
\label{sec:compression-factor}
To test whether our MACP model can handle compression data with minimum performance loss in real-world use cases, we examine the performance sensitivity against the compression factor on the V2V4Real Dataset. The compression factor is used in the cooperation ConAda
\begin{equation}
    \label{eq:compression-factor}
    \text{Compression Factor} = {C_\text{in}}/{C_\text{out}},
\end{equation}
where $C_\text{in}$ and $C_\text{out}$ are the number of input and output feature map channels of the down convolution. As shown in \cref{fig:compression}, the proposed MACP manages to maintain its performance and shows promising robustness against different compression rates.

\subsection{Cooperation Effectiveness}
\label{sec:cav-numbers}
While the MACP framework has shown promising performance, there is concern that it may not be able to utilize the extra information from V2V communication effectively. To test this, we conduct an experiment using OPV2V to examine whether increasing the number of CAVs can improve the model's performance. As shown in \cref{fig:num-cav}, the results indicate a clear positive correlation between prediction precision and the maximum number of CAV observations, demonstrating that our model can effectively make use of the extra information collected from the surrounding vehicles. \cref{fig:vis} also provides visualization results supporting this conclusion.

\section{Conclusion}
\label{sec:conclusion}
In this work, we proposed a novel framework to adapt single-agent models for cooperative perception efficiently. We addressed key challenges including domain shifts, computation, and communication constraints in real-world V2V applications. We have achieved superior performance in simulation-based and real-world cooperative perception benchmarks with high parameter efficiency and lower communication costs.
Our limitation is the assumption of ideal communication and localization. We'll cover transmission delay and other real-world factors in the future.



{\small
\bibliographystyle{bib/ieee_fullname}
\bibliography{bib/ma}

\begin{thebibliography}{10}\itemsep=-1pt

\bibitem{bai_transfusion_2022}
Xuyang Bai, Zeyu Hu, Xinge Zhu, Qingqiu Huang, Yilun Chen, Hongbo Fu, and
  Chiew-Lan Tai.
\newblock {TransFusion}: {Robust} {LiDAR}-{Camera} {Fusion} for {3D} {Object}
  {Detection} {With} {Transformers}.
\newblock In {\em {CVPR}}, pages 1090--1099, 2022.

\bibitem{bapna_simple_2019}
Ankur Bapna and Orhan Firat.
\newblock Simple, {Scalable} {Adaptation} for {Neural} {Machine} {Translation}.
\newblock In {\em {EMNLP}}, pages 1538--1548, 2019.

\bibitem{bilen_integrated_2016}
Hakan Bilen and Andrea Vedaldi.
\newblock Integrated perception with recurrent multi-task neural networks.
\newblock In {\em {NeurIPS}}, volume~29, 2016.

\bibitem{caesar_nuscenes_2020}
Holger Caesar, Varun Bankiti, Alex~H. Lang, Sourabh Vora, Venice~Erin Liong,
  Qiang Xu, Anush Krishnan, Yu Pan, Giancarlo Baldan, and Oscar Beijbom.
\newblock {nuScenes}: {A} {Multimodal} {Dataset} for {Autonomous} {Driving}.
\newblock In {\em {CVPR}}, pages 11621--11631, 2020.

\bibitem{caillot_survey_2022}
Antoine Caillot, Safa Ouerghi, Pascal Vasseur, Remi Boutteau, and Yohan Dupuis.
\newblock Survey on {Cooperative} {Perception} in an {Automotive} {Context}.
\newblock {\em IEEE Transactions on Intelligent Transportation Systems},
  23(9):14204--14223, 2022.

\bibitem{caruana_multitask_1997}
Rich Caruana.
\newblock Multitask {Learning}.
\newblock {\em Machine Learning}, 28(1):41--75, July 1997.

\bibitem{chen_f-cooper_2019}
Qi Chen, Xu Ma, Sihai Tang, Jingda Guo, Qing Yang, and Song Fu.
\newblock F-cooper: feature based cooperative perception for autonomous vehicle
  edge computing system using {3D} point clouds.
\newblock {\em ACM/IEEE Symposium on Edge Computing}, pages 88--100, 2019.

\bibitem{chen_cooper_2019}
Qi Chen, Sihai Tang, Qing Yang, and Song Fu.
\newblock Cooper: {Cooperative} {Perception} for {Connected} {Autonomous}
  {Vehicles} {Based} on {3D} {Point} {Clouds}.
\newblock {\em IEEE International Conference on Distributed Computing Systems
  (ICDCS)}, pages 514--524, 2019.

\bibitem{cui_radar_2023}
Can Cui, Yunsheng Ma, Juanwu Lu, and Ziran Wang.
\newblock Radar {Enlighten} the {Dark}: {Enhancing} {Low}-{Visibility}
  {Perception} for {Automated} {Vehicles} with {Camera}-{Radar} {Fusion}.
\newblock In {\em {IEEE} {International} {Conference} on {Intelligent}
  {Transportation} {Systems} ({ITSC})}, 2023.

\bibitem{deng_voxel_2021}
Jiajun Deng, Shaoshuai Shi, Peiwei Li, Wengang Zhou, Yanyong Zhang, and
  Houqiang Li.
\newblock Voxel {R}-{CNN}: {Towards} {High} {Performance} {Voxel}-based {3D}
  {Object} {Detection}.
\newblock In {\em {AAAI}}, volume~35, pages 1201--1209, 2021.

\bibitem{desislavov_trends_2023}
Radosvet Desislavov, Fernando Martínez-Plumed, and José Hernández-Orallo.
\newblock Trends in {AI} inference energy consumption: {Beyond} the
  performance-vs-parameter laws of deep learning.
\newblock {\em Sustainable Computing: Informatics and Systems}, 38:100857, Apr.
  2023.

\bibitem{dosovitskiy_carla_2017}
Alexey Dosovitskiy, German Ros, Felipe Codevilla, Antonio Lopez, and Vladlen
  Koltun.
\newblock {CARLA}: {An} {Open} {Urban} {Driving} {Simulator}.
\newblock In {\em {CoRL}}, volume~78, pages 1--16, 2017.

\bibitem{duan_centernet_2019}
Kaiwen Duan, Song Bai, Lingxi Xie, Honggang Qi, Qingming Huang, and Qi Tian.
\newblock {CenterNet}: {Keypoint} {Triplets} for {Object} {Detection}.
\newblock In {\em {ICCV}}, pages 6569--6578, 2019.

\bibitem{graham_3d_2018}
Benjamin Graham, Martin Engelcke, and Laurens van~der Maaten.
\newblock {3D} {Semantic} {Segmentation} {With} {Submanifold} {Sparse}
  {Convolutional} {Networks}.
\newblock In {\em {CVPR}}, pages 9224--9232, 2018.

\bibitem{graham_submanifold_2017}
Benjamin Graham and Laurens van~der Maaten.
\newblock Submanifold {Sparse} {Convolutional} {Networks}, June 2017.
\newblock arXiv:1706.01307 [cs.NE].

\bibitem{guo_parameter-efficient_2021}
Demi Guo, Alexander Rush, and Yoon Kim.
\newblock Parameter-{Efficient} {Transfer} {Learning} with {Diff} {Pruning}.
\newblock In {\em {ACL}}, pages 4884--4896, 2021.

\bibitem{he_deep_2016}
Kaiming He, Xiangyu Zhang, Shaoqing Ren, and Jian Sun.
\newblock Deep {Residual} {Learning} for {Image} {Recognition}.
\newblock In {\em {CVPR}}, pages 770--778, 2016.

\bibitem{hendrycks_gaussian_2016}
Dan Hendrycks and Kevin Gimpel.
\newblock Gaussian error linear units (gelus), 2016.
\newblock arXiv:1606.08415 [cs.LG].

\bibitem{houlsby_parameter-efficient_2019}
Neil Houlsby, Andrei Giurgiu, Stanislaw Jastrzebski, Bruna Morrone, Quentin
  De~Laroussilhe, Andrea Gesmundo, Mona Attariyan, and Sylvain Gelly.
\newblock Parameter-{Efficient} {Transfer} {Learning} for {NLP}.
\newblock In {\em {ICML}}, volume~97, June 2019.

\bibitem{hu_lora_2022}
Edward~J. Hu, Yelong Shen, Phillip Wallis, Zeyuan Allen-Zhu, Yuanzhi Li, Shean
  Wang, Lu Wang, and Weizhu Chen.
\newblock {LoRA}: {Low}-{Rank} {Adaptation} of {Large} {Language} {Models}.
\newblock In {\em {ICLR}}, 2022.

\bibitem{kirkpatrick_overcoming_2017}
James Kirkpatrick, Razvan Pascanu, Neil Rabinowitz, Joel Veness, Guillaume
  Desjardins, Andrei~A. Rusu, Kieran Milan, John Quan, Tiago Ramalho, Agnieszka
  Grabska-Barwinska, Demis Hassabis, Claudia Clopath, Dharshan Kumaran, and
  Raia Hadsell.
\newblock Overcoming catastrophic forgetting in neural networks.
\newblock {\em Proceedings of the National Academy of Sciences},
  114(13):3521--3526, Mar. 2017.

\bibitem{koh_wilds_2021}
Pang~Wei Koh, Shiori Sagawa, Henrik Marklund, Sang~Michael Xie, Marvin Zhang,
  Akshay Balsubramani, Weihua Hu, Michihiro Yasunaga, Richard~Lanas Phillips,
  Irena Gao, Tony Lee, Etienne David, Ian Stavness, Wei Guo, Berton Earnshaw,
  Imran Haque, Sara~M. Beery, Jure Leskovec, Anshul Kundaje, Emma Pierson,
  Sergey Levine, Chelsea Finn, and Percy Liang.
\newblock {WILDS}: {A} {Benchmark} of in-the-{Wild} {Distribution} {Shifts}.
\newblock In {\em {ICML}}, pages 5637--5664, 2021.

\bibitem{lang_pointpillars_2019}
Alex~H. Lang, Sourabh Vora, Holger Caesar, Lubing Zhou, Jiong Yang, and Oscar
  Beijbom.
\newblock {PointPillars}: {Fast} {Encoders} for {Object} {Detection} {From}
  {Point} {Clouds}.
\newblock In {\em {CVPR}}, pages 12697--12705, 2019.

\bibitem{li_learning_2021}
Yiming Li, Shunli Ren, Pengxiang Wu, Siheng Chen, Chen Feng, and Wenjun Zhang.
\newblock Learning {Distilled} {Collaboration} {Graph} for {Multi}-{Agent}
  {Perception}.
\newblock In {\em {NeurIPS}}, volume~34, pages 29541--29552, 2021.

\bibitem{li_learning_2018}
Zhizhong Li, Derek Hoiem, Andrei~A. Rusu, Neil~C. Rabinowitz, Guillaume
  Desjardins, Hubert Soyer, James Kirkpatrick, Koray Kavukcuoglu, Razvan
  Pascanu, and Raia Hadsell.
\newblock Learning without {Forgetting}.
\newblock {\em IEEE Transactions on Pattern Analysis and Machine Intelligence},
  40(12):2935--2947, Dec. 2018.

\bibitem{lian_scaling_2022}
Dongze Lian, Daquan Zhou, Jiashi Feng, and Xinchao Wang.
\newblock Scaling \& {Shifting} {Your} {Features}: {A} {New} {Baseline} for
  {Efficient} {Model} {Tuning}.
\newblock In {\em {NeurIPS}}, volume~35, pages 109--123, Dec. 2022.

\bibitem{liu_bevfusion_2023}
Zhijian Liu, Haotian Tang, Alexander Amini, Xinyu Yang, Huizi Mao, Daniela Rus,
  and Song Han.
\newblock {BEVFusion}: {Multi}-{Task} {Multi}-{Sensor} {Fusion} with {Unified}
  {Bird}'s-{Eye} {View} {Representation}.
\newblock In {\em {ICRA}}, 2023.

\bibitem{loshchilov_decoupled_2019}
Ilya Loshchilov and Frank Hutter.
\newblock Decoupled {Weight} {Decay} {Regularization}.
\newblock In {\em {ICLR}}, 2019.

\bibitem{marvasti_cooperative_2020}
Ehsan~Emad Marvasti, Arash Raftari, Amir~Emad Marvasti, Yaser~P. Fallah, Rui
  Guo, and Hongsheng Lu.
\newblock Cooperative {LIDAR} {Object} {Detection} via {Feature} {Sharing} in
  {Deep} {Networks}.
\newblock In {\em {VTC}}, pages 1--7, 2020.

\bibitem{qi_pointnet_2017}
Charles~Ruizhongtai Qi, Li Yi, Hao Su, and Leonidas~J Guibas.
\newblock {PointNet}++: {Deep} {Hierarchical} {Feature} {Learning} on {Point}
  {Sets} in a {Metric} {Space}.
\newblock In {\em {NeurIPS}}, volume~30, 2017.

\bibitem{qiao_adaptive_2023}
Donghao Qiao and Farhana Zulkernine.
\newblock Adaptive {Feature} {Fusion} for {Cooperative} {Perception} {Using}
  {LiDAR} {Point} {Clouds}.
\newblock In {\em {WACV}}, pages 1186--1195, 2023.

\bibitem{rauch_car2x-based_2012}
Andreas Rauch, Felix Klanner, Ralph Rasshofer, and Klaus Dietmayer.
\newblock {Car2X}-based perception in a high-level fusion architecture for
  cooperative perception systems.
\newblock In {\em {IEEE} {IV}}, pages 270--275, 2012.

\bibitem{rebuffi_learning_2017}
Sylvestre-Alvise Rebuffi, Hakan Bilen, and Andrea Vedaldi.
\newblock Learning multiple visual domains with residual adapters.
\newblock In {\em {NeurIPS}}, volume~30, 2017.

\bibitem{rockl_v2v_2008}
Matthias Rockl, Thomas Strang, and Matthias Kranz.
\newblock {V2V} {Communications} in {Automotive} {Multi}-{Sensor}
  {Multi}-{Target} {Tracking}.
\newblock In {\em {VTC}}, pages 1--5, 2008.

\bibitem{rusu_progressive_2018}
Andrei~A. Rusu, Neil~C. Rabinowitz, Guillaume Desjardins, Hubert Soyer, James
  Kirkpatrick, Koray Kavukcuoglu, Razvan Pascanu, and Raia Hadsell.
\newblock Progressive {Neural} {Networks}, Dec. 2018.
\newblock arXiv:1606.04671 [cs].

\bibitem{sugiyama_covariate_2007}
Masashi Sugiyama, Matthias Krauledat, and Klaus-Robert Müller.
\newblock Covariate {Shift} {Adaptation} by {Importance} {Weighted} {Cross}
  {Validation}.
\newblock {\em The Journal of Machine Learning Research}, 8:985--1005, 2007.

\bibitem{sung_training_2021}
Yi-Lin Sung, Varun Nair, and Colin~A Raffel.
\newblock Training {Neural} {Networks} with {Fixed} {Sparse} {Masks}.
\newblock In {\em {NeurIPS}}, volume~34, pages 24193--24205, 2021.

\bibitem{vaswani_attention_2017}
Ashish Vaswani, Noam Shazeer, Niki Parmar, Jakob Uszkoreit, Llion Jones,
  Aidan~N. Gomez, Lukasz Kaiser, and Illia Polosukhin.
\newblock Attention is all you need.
\newblock In {\em {NeurIPS}}, volume~30, 2017.

\bibitem{wang_v2vnet_2020}
Tsun-Hsuan Wang, Sivabalan Manivasagam, Ming Liang, Bin Yang, Wenyuan Zeng, and
  Raquel Urtasun.
\newblock {V2VNet}: {Vehicle}-to-{Vehicle} {Communication} for {Joint}
  {Perception} and {Prediction}.
\newblock In {\em {ECCV}}, pages 605--621, 2020.

\bibitem{xiang_hm-vit_2023}
Hao Xiang, Runsheng Xu, and Jiaqi Ma.
\newblock {HM}-{ViT}: {Hetero}-modal {Vehicle}-to-{Vehicle} {Cooperative}
  perception with vision transformer, Apr. 2023.
\newblock arXiv:2304.10628 [cs].

\bibitem{xu_pointfusion_2018}
Danfei Xu, Dragomir Anguelov, and Ashesh Jain.
\newblock {PointFusion}: {Deep} {Sensor} {Fusion} for {3D} {Bounding} {Box}
  {Estimation}.
\newblock In {\em {CVPR}}, pages 244--253, 2018.

\bibitem{xu_cobevt_2022}
Runsheng Xu, Zhengzhong Tu, Hao Xiang, Wei Shao, Bolei Zhou, and Jiaqi Ma.
\newblock {CoBEVT}: {Cooperative} {Bird}’s {Eye} {View} {Semantic}
  {Segmentation} with {Sparse} {Transformers}.
\newblock In {\em {CoRL}}, 2022.

\bibitem{xu_v2v4real_2023}
Runsheng Xu, Xin Xia, Jinlong Li, Hanzhao Li, Shuo Zhang, Zhengzhong Tu,
  Zonglin Meng, Hao Xiang, Xiaoyu Dong, Rui Song, Hongkai Yu, Bolei Zhou, and
  Jiaqi Ma.
\newblock {V2V4Real}: {A} {Real}-world {Large}-scale {Dataset} for
  {Vehicle}-to-{Vehicle} {Cooperative} {Perception}.
\newblock In {\em {CVPR}}, 2023.

\bibitem{xu_opencda_2023}
Runsheng Xu, Hao Xiang, Xu Han, Xin Xia, Zonglin Meng, Chia-Ju Chen, Camila
  Correa-Jullian, and Jiaqi Ma.
\newblock The {OpenCDA} {Open}-{Source} {Ecosystem} for {Cooperative} {Driving}
  {Automation} {Research}.
\newblock {\em IEEE Transactions on Intelligent Vehicles}, 8(4):2698--2711,
  Apr. 2023.

\bibitem{xu_v2x-vit_2022}
Runsheng Xu, Hao Xiang, Zhengzhong Tu, Xin Xia, Ming-Hsuan Yang, and Jiaqi Ma.
\newblock {V2X}-{ViT}: {Vehicle}-to-{Everything} {Cooperative} {Perception}
  with {Vision} {Transformer}.
\newblock In {\em {ECCV}}, pages 107--124, Aug. 2022.

\bibitem{xu_opv2v_2022}
Runsheng Xu, Hao Xiang, Xin Xia, Xu Han, Jinlong Li, and Jiaqi Ma.
\newblock {OPV2V}: {An} {Open} {Benchmark} {Dataset} and {Fusion} {Pipeline}
  for {Perception} with {Vehicle}-to-{Vehicle} {Communication}.
\newblock In {\em {ICRA}}, pages 2583--2589, 2022.

\bibitem{yan_second_2018}
Yan Yan, Yuxing Mao, and Bo Li.
\newblock {SECOND}: {Sparsely} {Embedded} {Convolutional} {Detection}.
\newblock {\em Sensors}, 18(10):3337, Oct. 2018.

\bibitem{yang_aim_2023}
Taojiannan Yang, Yi Zhu, Yusheng Xie, Aston Zhang, Chen Chen, and Mu Li.
\newblock {AIM}: {Adapting} {Image} {Models} for {Efficient} {Video} {Action}
  {Recognition}.
\newblock In {\em {ICLR}}, 2023.

\bibitem{yin_center-based_2021}
Tianwei Yin, Xingyi Zhou, and Philipp Krahenbuhl.
\newblock Center-{Based} {3D} {Object} {Detection} and {Tracking}.
\newblock In {\em {CVPR}}, pages 11784--11793, 2021.

\bibitem{yu_dair-v2x_2022}
Haibao Yu, Yizhen Luo, Mao Shu, Yiyi Huo, Zebang Yang, Yifeng Shi, Zhenglong
  Guo, Hanyu Li, Xing Hu, Jirui Yuan, and Zaiqing Nie.
\newblock {DAIR}-{V2X}: {A} {Large}-{Scale} {Dataset} for
  {Vehicle}-{Infrastructure} {Cooperative} {3D} {Object} {Detection}.
\newblock In {\em {CVPR}}, pages 21361--21370, 2022.

\bibitem{zhang_robust_2013}
Tianzhu Zhang, Bernard Ghanem, Si Liu, and Narendra Ahuja.
\newblock Robust {Visual} {Tracking} via {Structured} {Multi}-{Task} {Sparse}
  {Learning}.
\newblock {\em IJCV}, 101(2):367--383, 2013.

\bibitem{zhang_facial_2014}
Zhanpeng Zhang, Ping Luo, Chen~Change Loy, and Xiaoou Tang.
\newblock Facial {Landmark} {Detection} by {Deep} {Multi}-task {Learning}.
\newblock In {\em {ECCV}}, pages 94--108, 2014.

\bibitem{zhang_distributed_2021}
Zijian Zhang, Shuai Wang, Yuncong Hong, Liangkai Zhou, and Qi Hao.
\newblock Distributed {Dynamic} {Map} {Fusion} via {Federated} {Learning} for
  {Intelligent} {Networked} {Vehicles}.
\newblock In {\em {ICRA}}, pages 953--959, 2021.

\bibitem{zhao_end--end_2020}
Sicheng Zhao, Yunsheng Ma, Yang Gu, Jufeng Yang, Tengfei Xing, Pengfei Xu,
  Runbo Hu, Hua Chai, and Kurt Keutzer.
\newblock An {End}-to-{End} {Visual}-{Audio} {Attention} {Network} for
  {Emotion} {Recognition} in {User}-{Generated} {Videos}.
\newblock In {\em {AAAI}}, volume~34, pages 303--311, 2020.

\bibitem{zhao_epointda_2021}
Sicheng Zhao, Yezhen Wang, Bo Li, Bichen Wu, Yang Gao, Pengfei Xu, Trevor
  Darrell, and Kurt Keutzer.
\newblock {ePointDA}: {An} {End}-to-{End} {Simulation}-to-{Real} {Domain}
  {Adaptation} {Framework} for {LiDAR} {Point} {Cloud} {Segmentation}.
\newblock In {\em {AAAI}}, volume~35, pages 3500--3509, 2021.

\bibitem{zhou_voxelnet_2018}
Yin Zhou and Oncel Tuzel.
\newblock {VoxelNet}: {End}-to-{End} {Learning} for {Point} {Cloud} {Based}
  {3D} {Object} {Detection}.
\newblock In {\em {CVPR}}, pages 4490--4499, 2018.

\bibitem{zhou_centerformer_2022}
Zixiang Zhou, Xiangchen Zhao, Yu Wang, Panqu Wang, and Hassan Foroosh.
\newblock {CenterFormer}: {Center}-{Based} {Transformer} for {3D} {Object}
  {Detection}.
\newblock In {\em {ECCV}}, pages 496--513, 2022.

\end{thebibliography}
}

\appendix

\section{Implementation Details}
\subsection{Model}
In our implementation, we use a weighted sum strategy to combine the feature maps of the ego vehicle and the surrounding vehicles. The weight of the ego vehicle feature map is set to $1$, while the weights of the surrounding vehicles are set to $1/N$, where $N$ is the number of surrounding vehicles.

After combining all feature maps from the ego and surrounding vehicles, we add another $3\times3$ convolution layer (with normalization and activation layers) to further adjust for any spatial misalignment, following the approach of BEVFusion~\cite{liu_bevfusion_2023}.

We adopt the center-based strategy to predict the locations of objects and employ several regression heads to estimate object size and heading. For further details, please refer to previous studies on 3D object detection~\cite{yin_center-based_2021,bai_transfusion_2022}.

\subsection{Training}
The voxel size is set to $(0.200, 0.075, 0.200)$ for the $x$, $y$, and $z$ directions, respectively. We apply common point cloud data augmentation techniques to prevent overfitting, such as scaling and rotation~\cite{lang_pointpillars_2019}. The learning rate is initially set to $2\times10^{-5}$, and we schedule the learning rate with a cosine annealing. For the main results, we trained our models for 20 epochs with a batch size of 4 per GPU (NVIDIA A100).

\section{Discussion}
\paragraph{Different Fusion Methods}
We conduct an ablation study to compare the weighted sum method with other feature fusion strategies. These strategies include sum (directly adding up the feature maps), mean (calculating the average of all feature maps), and concat (computing the mean of the feature maps of the surrounding vehicles and then concatenating it with the feature map of the ego vehicle, doubling the number of channels). All the methods are compared with a compression factor of 4, and the models are trained for 5 epochs on the OPV2V dataset~\cite{xu_opv2v_2022}. As seen in \cref{tab:fusion}, the weighted sum approach achieves the top overall performance.

\begin{table}[!th]
    \begin{center}
        \resizebox{\linewidth}{!}{
            \begin{tabular}{l|cccc}
                \hline
                \multirow{2}{3em}{Method}  &\multicolumn{4}{c}{AP@IoU=50/70 $(\uparrow)$}\\
                & Overall & 0-30m & 30-50m & 50-100m \\
                \hline
                Mean & 89.6/86.2 & 97.7/96.3 & 92.2/88.0 & 77.7/72.5\\
                Sum & 74.8/68.9 & 91.1/87.1 & 72.0/64.6 & 54.4/47.7\\
                Concat & 91.9/86.2 & \textbf{98.0}/96.3 & \textbf{92.2}/\textbf{86.7} & 84.5/74.1\\
                Weighted Sum (\textbf{MACP}) & \textbf{92.4}/\textbf{87.1} & \textbf{98.0}/\textbf{96.6} & 92.0/86.3 & \textbf{85.9}/\textbf{76.5}\\
                \hline
            \end{tabular}
        }
    \end{center}
    \caption{\textbf{Ablation Results on Fusion Methods.}}
    \label{tab:fusion}
\end{table}

\begin{figure}[!th]
    \centering
    \includegraphics[width=\linewidth]{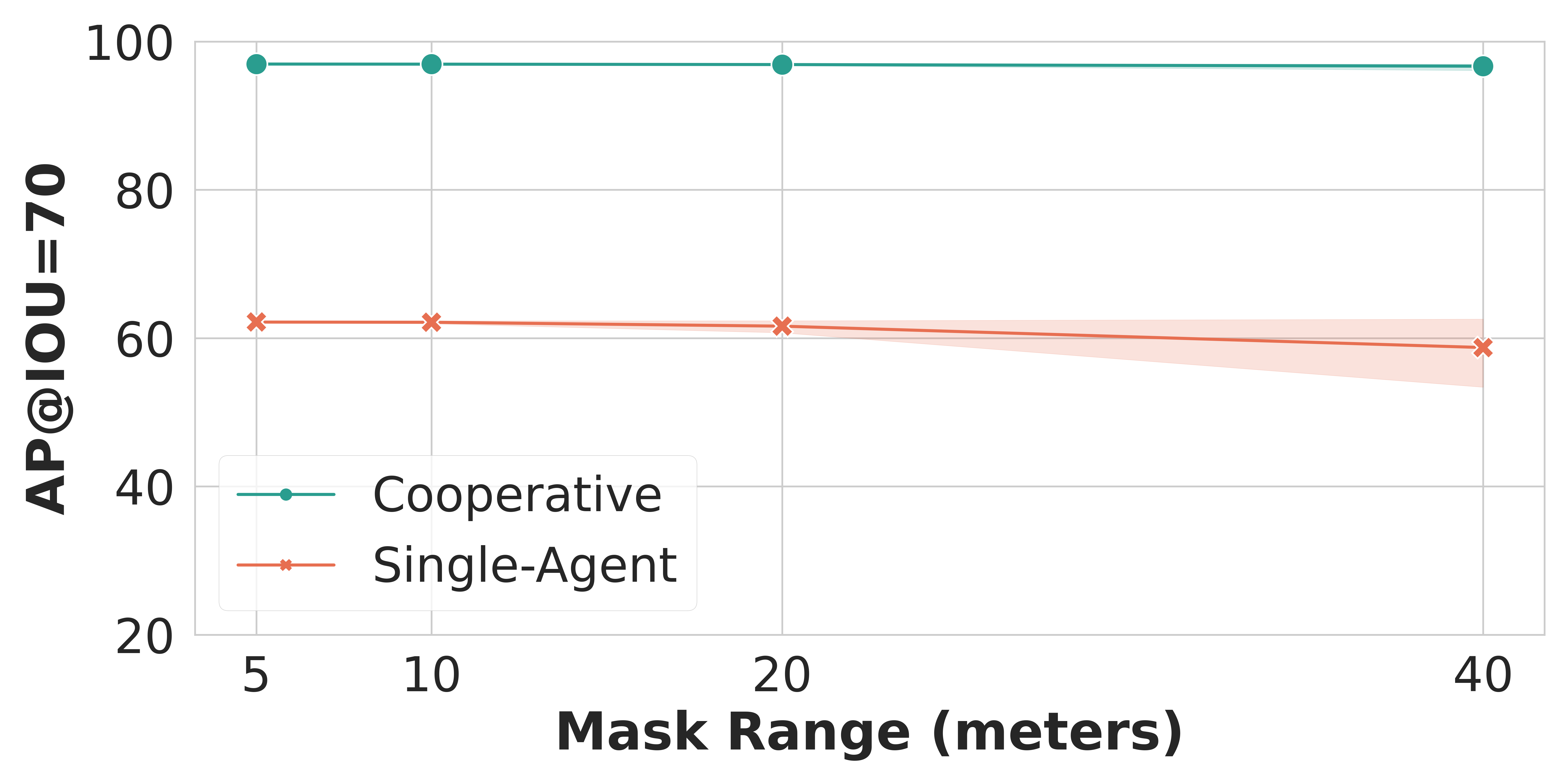}
    \caption{\textbf{Results of the robustness analysis.} Visualization reflects that the extra information from cooperative perception significantly helps the model tackle the drop in performance due to occlusions or other factors causing missing observations.}
    \label{fig:robust}
\end{figure}

\paragraph{Robustness}
Taking advantage of an extended field of view from V2V cooperative perception, the proposed MACP model should be able to tackle the occlusion or sudden decreases in visibility, that is, to have higher robustness. We design an experiment to assess the robustness by traversing and masking out part of the ego vehicle's field of view and looking into how it affects performance. \cref{fig:robust} visualizes the results, where with an occlusion range of 40 by 40 meters, the single-agent perception model struggles to maintain a stable performance, revealed by an increasing variance in average precision. On the contrary, the cooperative perception model consistently retains its prediction accuracy, exhibiting an AP standard deviation of 1.08\%, demonstrating remarkable robustness.

\end{document}